\documentclass[letterpaper]{article}
\usepackage{amsmath,graphicx,mlspconf}
\usepackage{multirow}
\usepackage{color}
\usepackage{amssymb}
\usepackage{url}
\usepackage{setspace}
\copyrightnotice{978-1-5090-6341-3/17/\$31.00 {\copyright}2017 IEEE}

\toappear{2017 IEEE International Workshop on Machine Learning for Signal Processing, Sept.\ 25--28, 2017, Tokyo, Japan}

\def\L{{\cal L}}

\newcommand{\compresslist}{%
\setlength{\itemsep}{1pt}%
\setlength{\parskip}{0pt}%
\setlength{\parsep}{0pt}%
}

\title{Deep convolutional neural networks for \\interpretable analysis of EEG sleep stage scoring}
\name{Albert Vilamala$^{1}$, Kristoffer H. Madsen$^{1,2}$ and Lars K. Hansen$^{1}$\thanks{This project has received funding from the European Union's Horizon 2020 research and innovation programme under the Marie Sklodowska-Curie grant agreement No 659860. We gratefully acknowledge the support of NVIDIA Corporation with the donation of the GPUs used for this research.}}
\address{
\begin{tabular}{c c}
  $^1$Technical University of Denmark & $^2$Danish Research Centre for Magnetic Resonance \\
\{alvmu, lkai\}@dtu.dk & kristofferm@drcmr.dk
\relax
 \end{tabular}
}
\begin{document}
%\ninept
%

%{ {$^2$Danish Research Centre for Magnetic Resonance\\
% kristofferm@drcmr.dk}}
 
\maketitle
\begin{abstract}
%\textcolor{red}{[100 - 150 words] Review abstract}
Sleep studies are important for diagnosing sleep disorders such as insomnia, narcolepsy or sleep apnea. They rely on manual scoring of sleep stages from raw polisomnography signals, which is a tedious visual task requiring the workload of highly trained professionals. Consequently, research efforts to purse for an automatic stage scoring based on machine learning techniques have been carried out over the last years. 
In this work, we resort to multitaper spectral analysis to create visually interpretable images of sleep patterns from EEG signals as inputs to a deep convolutional network trained to solve visual recognition tasks. As a working example of transfer learning, a system able to accurately classify sleep stages in new unseen patients is presented. Evaluations in a widely-used publicly available dataset favourably compare to state-of-the-art results, while providing a framework for visual interpretation of outcomes.
\end{abstract}

%%%%% Introduction %%%%%
%\section{Introduction}
%\label{sec:intro}
\section{Sleep stage scoring}
\label{sec:sleep}
Studies of sleep assist doctors to diagnose sleep disorders and provide the baseline for appropriate follow up. Clinical sleep study design is based on polysomnography (PSG) in which several  biological signals are acquired while the patient is asleep, including electroencephalography (EEG) for monitoring brain activity, electrooculogram (EOG) for eye movements and electromyogram (EMG) to measure muscle tone. The measurements are used to classify sleep score, i.e., classify the sleep stages the patient goes through during the study and assess the existence of any dysfunction. With the increasing accessibility of EEG signals (e.g., using permanent implanted electrodes or semi-permanent measures through sensors placed, e.g., in the ear \cite{Mikkelsen2015}) there is a growing interest in sleep quantification based on EEG alone. The EEG signal often presents significant bursts of rhythmic  components, In particular the frequency band `alpha' (8-12 Hz) is prominent. The American Academy of Sleep Medicine (AASM) recommends segmentation of sleep in five stages \cite{Berry2017}:
\begin{itemize}\compresslist
\item W (wakefulness): alpha  (8-12 Hz) rhythm is present, high-amplitude muscle contractions and movement artefacts on EMG and eye blinking on EOG, which can also be appreciated in low frequency EEG (0.5-2 Hz). 
\item N1 (Non-REM 1): alpha (8-12 Hz) rhythm is attenuated and replaced by mixed frequency theta signal (4-7 Hz), decrease in muscle tone and slow eye movements.
\item N2 (Non-REM 2): presents K-complexes (negative peak followed by a positive complex and a final negative voltage) in the $<$1.5 Hz range and sleep spindles (burst of oscillatory waves) at sigma (12-15 Hz) band.
\item N3 (Non-REM 3): slow wave activity exists (0.5-3 Hz), eye movements are unusual and EMG tone is low.
\item R (REM): existence of rapid eye movements clearly visible in EOG, relatively low-amplitude and mixed-frequency activity in EEG, and presents the lowest muscle tone on EMG. %\textcolor{red}{I think there is also some theta activity in REM?}
\end{itemize}
Manually scoring sleep stages is a tedious task requiring sleep experts to \emph{visually} inspect PSG data recorded during the whole sleep study. Thus, there has been considerable effort over the past years to develop machine learning (ML) methods for automatic sleep scoring. For a recent survey on the available literature refer to \cite{Aboalayon2016}. Recent research based on fully-connected artificial neural networks (ANN) with a variety of hand-crafted features extracted from the EEG signal includes \cite{Ronzhina2012, Ebrahimi2008}. More recently,  recurrent neural network variants have been used to capture long-term dependencies or stage transition rules \cite{Hsu2013,Supratak2017}.

Here our aim is to discuss and test the hypothesis that sleep stage classification can be assisted by \emph{transfer learning} (e.g., support the classifier training process by use of related data). We will take advantage of the fact that sleep scoring is a {\it visual} classification problem, hence, the strong progress in developing artificial visual perception using convolutive neural networks (CNNs, \cite{LeCun1989,Krizhevsky2012,Simonyan2014,He2016}). In particular, time-frequency spectrogram images are created from windowed EEG signals and fed to a CNN pre-trained on a visual object recognition task, allowing the use of this powerful model for sleep stage classification in EEG data, which would otherwise lead to overfitting due to limited availability of data.

Despite the quite extensive research efforts in sleep scoring, there exist several challenges related to evaluation and comparison: one is the large variety of datasets (most outside the public domain) used to evaluate the methods; another frequent obstacle is the design of experiments and evaluation procedures, specifically proper cross-validation procedures. Some use a single train/test split, where no cross-validation is performed; others completely lack an independent test set, hence have no unbiased performance measure. Finally, others neglect the important dependency structures in the data when cross-validating, e.g., by completely random sampling. Random sampling ignores the strong dependence between data from the same subject or even obtained on the same night.

In this paper we have selected two recent results for comparison of performance. They both have been cross-validated in a proper sense and provide relevant information for reproducibility. In \cite{Tsinalis2016a}, authors conduct a time-frequency analysis of the EEG signal to extract relevant features to feed an ensemble of stacked sparse autoencoders. In a later investigation \cite{Tsinalis2016b}, authors construct an end-to-end ANN by combining a CNN architecture using raw EEG signals with a 2D stack of frequency-specific activity over time. Both studies obtain state-of-the-art results.

%An important aspect to take into consideration when implementing ANN for the real use in certain domains (e.g. in the medical field) is precisely that of understanding the internal functioning of the model

ANNs have often been criticized to work as black boxes: input data are fed into the one end and output values are obtained at the other end. Understanding the complex internal transformations that ANNs perform to the data is crucial for both domain practitioners, who need to comprehend the ANNs functioning in order to rely on them, and ML engineers, who require a deep knowledge of the system for correcting errors and improving the models. Therefore, different efforts have been made to provide a set of visualisation techniques to aid on this matter \cite{Simonyan2013,Zeiler2014}.

Our contributions in this paper are the following: we provide a framework for the automatic analysis of single sensor EEG sleep stage scoring, which is both highly accurate and visually interpretable by first using multitaper spectral estimation to generate colour image spectrograms, containing natural image-like features (blobs, vertices, etc.), emphasizing sleep patterns for the experts to analyse; secondly, given the natural image look of our inputs, we employ and refine the super-accurate models trained on natural images, which have been proven to excel on natural image visualisation, to classify the different sleep stages; and finally, we make use of sensitivity analysis to map the most influential features in our network back to the input space, providing highly interpretable images about the network's functioning. In particular, we employ a highly used publicly available database to classify the $5$ sleep stages mentioned before with an average accuracy of $86\%$ on an independent test set, but more importantly, we report the time-frequencies of interest for our network to decide on each sleep stage.

%Focusing the attention to our domain of applicability, there have been many studies using ANNs to analyse EEG signals (\textcolor{red}{cite some works}), but to the best of our knowledge just few of them have treated the problem as a visualisation task using CNNs (\textcolor{red}{check for other works on CNNs for EEG}). The most prominent one is the work in \cite{Bashivan2015}, where they created a set of topology-preserving multi-spectral images out of multi-channel EEG time-series and trained a deep recurrent CNN in order to find appropriate representations to analyse EEG data. The system was evaluated in a cognitive load classification task with improved results over state-of-the-art approaches.

%\subsection{Literature review}
%\label{ssec:sleep_literature}

%Electroencephalography (EEG) is a non-invasive technique that measures electrical activity in the brain. Due to its portability, affordability and temporal resolution, it has %become one of the preferred candidate tools to measure brain activity in the wild. However, the fact that the recording sensors are placed in the scalp and not in the internal %brain structures, from which the currents originate, renders the obtained signal very weak, noisy and heavily correlated across sensors. Therefore, it is often required to use %signal processing and machine learning (ML) techniques to extract relevant information from the signal and find patterns that help us understand the internal processes of the brain %when performing different tasks.

%%%%% Deep Neural Networks %%%%%%
\section{Transfer learning with convolutional neural networks}
\label{sec:deep_nets}
CNNs were originally introduced in the late 80's \cite{LeCun1989} as biologically inspired models to perform image recognition. However, a major breakthrough in the field occurred in 2012, when the deep CNN \emph{AlexNet} \cite{Krizhevsky2012}, won the \emph{ImageNet Large-Scale Visual Recognition Challenge} (ILSVRC) by a big margin to all other competing models. Since then, CNNs have become the workhorse for visualisation tasks and ILSVRC the preferred benchmark to test their performance. AlexNet consisted of five convolutional layers, max-pooling layers and three fully-connected layers. In 2013, ZFNet \cite{Zeiler2014} won the competition with a similar architecture to AlexNet but better optimised hyperparameters; authors also provided an approach to visualise the working of the CNN. GoogLeNet \cite{Szegedy2015} won the competition in 2014 with a CNN composed of 22 layers; many of which contained an \emph{inception module} with several convolutions and pooling in parallel, as well as replacing fully-connected layers by average-pooling. Nevertheless, a highly successful model this year was the VGGNet \cite{Simonyan2014}, thanks to its simplicity: it is a 16-layer network composed exclusively of $3 \times 3$ convolutions, $2 \times 2$ max-pooling and 3 fully-connected layers. In 2015, the challenge was won by ResNet \cite{He2016}, a 152-layer CNN that uses \emph{residual blocks} in which convolutions and activations do not compute a transformation function of the feature maps but rather the deviation from the identity function.

One big challenge when training such flexible models is to avoid overfitting: all previously mentioned CNNs use large amounts of training data and apply data augmentation and dropout techniques to address this issue. Still, there exist many domains in which acquiring input data is very costly (e.g., medical sciences) and training a deep neural network de novo is simply not feasible. 

Therefore \emph{transfer learning} is of interest. In the case of CNNs trained on natural images there seems to be a consistent behaviour in which lower layers of the network hierarchy learn general features similar to Gabor filters and colour blobs, while higher layers capture more domain specific representations. This observation has been studied in \cite{Yosinski2014}, where the layers' generality-specificity trade-offs have been quantitatively evaluated. Authors also analysed the decrease of transferability performance as the target task differs from the original one. In a different study \cite{Razavian2014}, authors use the features learned in an object recognition task to extract features from a varying set of visual problems and datasets, obtaining superior results than task-specific state-of-the-art systems. Finally, \cite{Donahue2013} used transfer learning in CNNs to extract features from unrelated tasks presenting insufficient data, an approach allowing the use of these flexible models.

%%%%%% Model %%%%%%%%%%%
\section{Methods}
\label{sec:model}
\subsection{Image creation}
\label{ssec:image_creation}
EEG data is represented as time-frequency images. Spectral estimation is based on Fourier analysis, which assumes a series of properties on the data such as infinite signals, continuity, periodicity and stationarity. However, none of these assumptions are typically fulfilled in EEG data, therefore in practise the spectrogram becomes a highly biased estimator. A common strategy to reduce the spectrogram's bias is by convolving the raw signal with a window function (called taper) before performing spectral estimation. Another important drawback of the spectrogram is that it produces estimates with high variance across all frequencies, which are even increased with the use of tapers. A method called \emph{multitaper spectral estimation} \cite{Thomson1982} can be used to reduce both bias and variance by applying multiple taper functions to the raw signal and averaging their results. %Importantly, tapers are selected from a class of orthogonal functions called \emph{discrete prolate spheroidal sequence} (DPSS). 
This technique has recently been proven to be highly successful in analysing neurophysiological dynamics of sleep using EEG \cite{Prerau2016}. %by overcoming ...
%
%?dont really understand the infinite signals assmptions
%
There exists a set of hyperparameters that highly influence the results of the multitaper spectral estimation: they are the window size $\omega$ in seconds, the window stepsize $\sigma$ in seconds, the minimum frequency resolution that can be resolved $f$ in Hz, the time-half-bandwidth product, usually defined as $W=\omega f/2$ and the number of tapers used, often set according to the heuristic $L = \lfloor2W \rfloor- 1$, where $\lfloor x \rfloor$ is a function that rounds $x$ down to the closest integer \cite{Prerau2016}.
After estimating the multitaper spectrogram, we convert its values to a logarithmic scale $x = \log(x) + 1$ and split it into equally-length bins of size $s$, called epochs. Then, we convert each epoched spectrogram to an RGB colour matrix (i.e., an image) by applying our preferred colourmap (i.e., a lookup table that translates real values in the $[0,1]$ interval to RGB colours).
%Then, we produce an image from each epoch by directly converting the spectrogram to an RGB colour matrix (i.e., an image) by applying our preferred colourmap (i.e., a lookup table that translates real values in the $[0,1]$ interval to RGB colours).
%
%To this point we have explained how to create an image from a single EEG sensor. \textcolor{red}{(Two sensors by mirroring the spectrograms? Or maybe creating an image out of each sensor and apply PCA afterwards.)} On the contrary, if we are interested in exactly $3$ sensors, we repeat the aforementioned procedure three times, up to the creation of the log-spectrogram matrix. Then, an image is obtained by storing each sensor-spectrogram as the R, G and B color channel. If we want to use more than $3$ channels, we first apply a decomposition technique to all sensors' time-series, such as Principal Components Analysis (PCA), and retrieve the first $3$ components as the pseudo-sensors. Now, we use the $3$-sensor procedure to create images using the pseudo-sensors, instead.
%
\subsection{Network architecture}
\label{ssec:network_architecture}
The CNN of choice to analyse the images created according to the previous block is the VGGNet \cite{Simonyan2014}, due to its simplicity and flexibility. It is composed of $16$ weighted layers:
\[
\mathrm{ccm_{64} ccm_{128} cccm_{256} cccm_{512}cccm_{512}fcr_{4096}fcr_{4096}fcs_{1000}},
\]
\normalsize
where $\mathrm{c}$ means a $3 \times 3$ convolutional filter of stride $1$ using a ReLU activation function, $\mathrm{m}$ stands for $2 \times 2$ max-pooling layer with a stride of $2$, $\mathrm{fcr}$ and $\mathrm{fcs}$ correspond to fully-connected layers with ReLU and soft-max activations, respectively; sub-indexed values represent the number of channels in each block. \emph{Transfer learning} is employed by using weight values in all convolutional layers that have been previously trained on ILSVRC-2014 data provided by the authors in \url{http://www.robots.ox.ac.uk/~vgg/research/very_deep}. Fully-connected layers are initialised from scratch using Xavier's initialisation \cite{Glorot2010} and trained using dropout. The number of final outputs is set according to the task we are tackling.
\subsection{Network visualisation}
\label{ssec:network_visualisation}
In the current study, we resort to sensitivity analysis \cite{Rasmussen2011, Hashem1992, Zurada1994} as a visualisation tool to better understand the decisions made by our network. More precisely, let $\mathcal{D}=\{\mathbf{x}_n, t_n\}_{n=1}^N$ be a dataset of $P$-dimensional input vectors $\mathbf{x}$ (i.e., spectral images in our work) and corresponding class labels $t \in \{1,\dots, C\}$, the built ANN acts as a function approximator, such that $\hat{t} = f(\mathbf{x})$. We can estimate the relative importance that our network places to every input feature $j$ (i.e., RGB colour channel in a pixel, in our context) to discriminate among the existing classes as:
\begin{equation}
\label{eq:sensitivity}
\hat{s}^{(j)} = \frac{1}{N}\sum_{n=1}^N \left| \frac{\partial \L(f(\mathbf{x}),t)}{\partial \mathbf{x}^{(j)}} \right|_{\mathbf{x}=\mathbf{x}_n}
\end{equation}
where $\L$ is the loss function of choice and $| x |$ is the absolute value of $x$. \emph{Sensitivity maps} are created by disposing $\hat{s}^{(j)}$ in the corresponding RGB colour matrix forming an image. The fact that most of the current frameworks supplying ANN building capabilities are provided with automatic differentiation procedures reduces the calculation of sensitivity maps to a simple function call.
%Notice that $\L(f(\mathbf{x}),t)$ introduced earlier evaluates the final output of the ANN; yet, nothing prevents this method to be used in intermediate layers of the network to visualise their specific functioning. 

%
%%%%% EEG Sleep scoring %%%%%%%%
\section{Empirical evaluation}
\label{sec:evaluation}
\subsection{Experimental setup}
\label{ssec:sleep_experiments}
\begin{table*}[tb]
\small
    \centering
  \begin{tabular}{l l | r r r r r | r r r r r | c  c  c  c}
   % \toprule
    & &\multicolumn{5}{c}{\bf{Predicted (aggregate)}} & \multicolumn{5}{c}{\bf{Normalised pred. (in \%)}} &\multicolumn{4}{c}{\bf{Per-class metrics (in \%)}}\\
    %\cmidrule{2-5}
    & & W & N1 & N2 & N3 & R & W & N1 & N2 & N3 & R &  Pre. & Sen. & F1 & Acc.\\
        %\midrule
    \hline
     \multirow{ 5}{*}{\bf{VGG-FE}} & W & 3529 & 579 & 97& 46 & 258 & 78 & 13 & 2 & 1 & 6 &  93 & 78 & 85 & 86 \\
     &N1 & 458 & 1219 & 353 & 29 & 703  & 17 & 44 & 13 & 1 & 25&  85 & 44 & 58 & 68 \\
     &N2 & 346 & 1215 & 13118 & 1676 & 1222 & 2 & 7 & 75 & 10 & 7 &  91 & 75 & 82 & 84 \\      
     &N3 & 80 & 31 & 461 & 5003 & 16 & 1 &  1 & 8 & 89 & 0 & 97 &  89 & 93 & 93 \\
     &R & 219 & 781 & 470 & 6 & 6235 & 3 & 10 & 6 & 0 & 81 &  89 & 81 & 85 & 86 \\
     \hline
     \multirow{ 5}{*}{\bf{VGG-FT}} & W & 3505 & 671 & 52 & 39 & 242 & 78 & 15 & 1 & 1 & 5 & 96 & 78 & 86 & 87 \\
     & N1 & 301 & 1553 & 334 & 19 & 555  & 11 & 56 & 12 & 1 & 20 & 89 & 56 & 69 & 75 \\
     & N2 & 192 & 985 & 13884 & 1411 & 1105 & 1 & 6 & 79 & 8 & 6  & 92 & 79 & 85 & 86 \\     
     & N3 & 73 & 24 & 462 & 5015 & 17  & 1 & 0 & 8 & 90 & 0 & 97 & 90 & 93 & 94 \\
     & R & 82 & 563 & 378 & 14 & 6674  & 1 & 7 & 5 & 0 & 87 & 92 & 87 & 89 & 89 \\
     \hline
    %\bottomrule
  \end{tabular}
  \caption{Raw confusion matrix, normalised confusion matrix and per-class metrics for both VGG-FE and VGG-FT.}
  \label{tab:confusion_matrix}
 %    \vspace{-5mm}
\end{table*}

We use EEG sleep recordings from the Sleep-EDF Database \cite{Kemp2000} in the PhysioNet repository \cite{Goldberger2000}. In particular, a subset of data from a study of age effects on sleep in healthy subjects, containing two whole-night EEG recordings (approximately 20 hours) from Fpz-Cz and Pz-Oz channels sampled at 100 Hz and corresponding hypnograms (expert annotations of sleep stages) from 20 subjects (10 males and 10 females) between 25-34 years old (second night of subject 13 was not provided). Sleeping time was retrieved from each recording as the interval between annotated lights off and lights on times or from 15 minutes before/after the first/last scored sleep epoch, if these annotations were not provided. Class labels were obtained from the hypnograms at every $30$ s.

Images were created for Fpz-Cz sensor as explained in Section~\ref{ssec:image_creation}, setting $\omega=3.0$ s., $f=2$ Hz, $W=3$ and $L=5$ tapers, with the purpose to capture the sleeping dynamics at the microevent time scale while maintaining a somewhat fine resolution \cite{Prerau2016}. The window stepsize was set to $\sigma=0.67$ s. in order to match the final image resolution (prefixed to $224 \times 224$ pixels by the pre-trained VGGNet). Bin size was set to $s=150$ s., corresponding to the current epoch plus the two previous and two posterior epochs, as it has been shown to improve overall accuracy by better classifying N1-N2, N1-R and N2-R transition stages \cite{Tsinalis2016a}. Spectrogram log values were thresholded to the $[0, 1]$ interval before applying the `Jet' colourmap to generate the images.

Following \cite{Tsinalis2016a, Tsinalis2016b}, we used a leave-one-subject-out sampling schema to partition the data into train and test datasets, further splitting the training dataset by randomly selecting $4$ subjects for validation and the remaining $15$ for actual training. The mentioned works address the skewed performance favouring the most represented classes during training in such imbalanced datasets using different approaches: first study employs class-balanced random sampling with an ensemble of classifiers, while the second one uses a different class-balanced batch at each stochastic gradient descent (SGD) epoch. We virtually use both strategies by randomly balancing all classes in each SGD epoch and obtaining ensemble behaviour thanks to the dropout layers \cite{Srivastava2014}.

CNNs were trained by optimising the categorical cross-entropy between predicted values and class labels using \emph{adam} \cite{Kingma2014} SGD on mini batches of 250 training examples with a learning rate of $10^{-5}$, and decay rate of first and second moments set to $0.9$ and $0.999$, respectively. The validation set was employed to choose the hyper-parameters and its loss as a stopping criterion to avoid overfitting.

\subsection{Results}
\label{ssec:sleep_results}
We evaluated our method in two different scenarios: first, we used the VGGNet as a feature extractor (VGG-FE), where all convolutional layers were kept fixed and only the last $3$ fully-connected layers of the network were trained from scratch to the specific EEG sleep scoring problem. In the second scenario, all weights in the network were updated, obtaining a fine-tuned network (VGG-FT). Preliminary results including a randomly initialised network consistently showed suboptimal classification accuracy, being it more pronounced the smaller the training set was. Convergence time for this network was on average 3 times slower than VGG-FT.

%As explained in Section~\ref{sec:deep_nets}, lower layers of the network are assumed to extract general low level features (e.g., lines, blobs), while higher levels of the network encode domain-specific high level representations.

We report the aggregated confusion matrix for both scenarios over all test sets (Table~\ref{tab:confusion_matrix}, left), out of which we compute all different scoring performance metrics. To do so, we first class-balance the confusion matrix (Table~\ref{tab:confusion_matrix}, middle) and split the results into $5$ binary one-vs-all classification, which are again renormalised in order to calculate the per-class \emph{precision, sensitivity, F1-score and accuracy} (Table~\ref{tab:confusion_matrix}, right). Comparison with existing literature is presented in Table~\ref{tab:literature_comparison}, where the 95\% confidence interval for each measure is computed using bootstrapping. In each of the 1000 bootstrap iterations, 20 confusion matrices out of the 20 subjects are sampled (with replacement) and their values aggregated. Then, each evaluation metric is calculated. The 1000 bootstrap samples per evaluation metric are ordered and the values at 26th, 500th and 975th positions are reported.

As shown in Table~\ref{tab:confusion_matrix}, top block, the most correctly sleep stage classified by VGG-FE is N3 with $89\%$ of epochs properly assigned. It is followed by R (81\%), W (78\%) and N2 (75\%). The most difficult stage to classify is N1, with $44\%$ of epochs correctly assigned. The highest misclassification rate for this stage is for assigning epochs to R stage (25\%), followed by W (17\%) and finally N2 (13\%); a behaviour that is consistent with the literature \cite{Tsinalis2016a,Tsinalis2016b}. These results seem to indicate that, while N1-N2 transitions have been acceptably captured by our method, N1-R clearly misses some important information. If we turn our attention towards per-class metrics (Table ~\ref{tab:confusion_matrix}, top-right), we can observe that their precisions range between $89-97\%$, except for the N1, which reaches a not inconsiderable $85\%$. However, sensitivities are scoring significantly lower, their values lying in the $75-89\%$ interval, with the exception of N1, which presents a score of $44\%$, highly impacting the accuracy and F1-score for this class.

\begin{figure*}[t]
\begin{minipage}[b]{0.19\linewidth}
  \centering
  \centerline{\includegraphics[width=\linewidth]{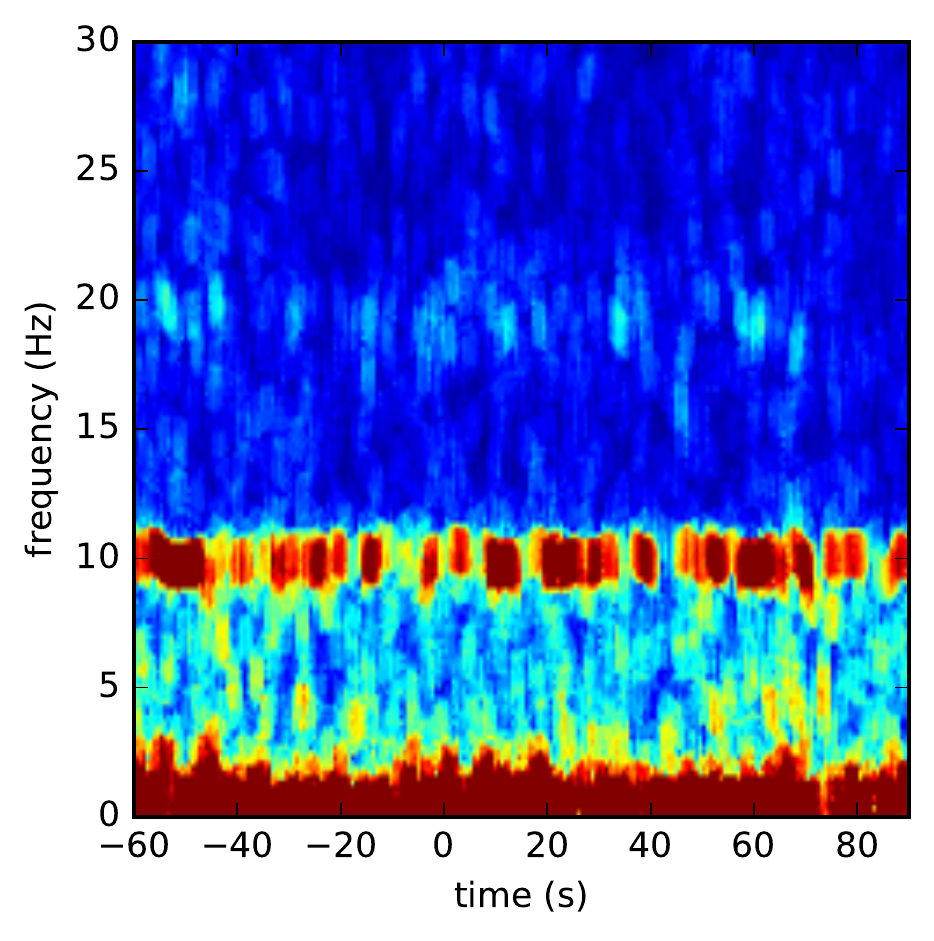}}
%  \vspace{2.0cm}
  \centerline{(a) Wakefulness}\medskip
\end{minipage}
\hfill
\begin{minipage}[b]{0.19\linewidth}
  \centering
  \centerline{\includegraphics[width=\linewidth]{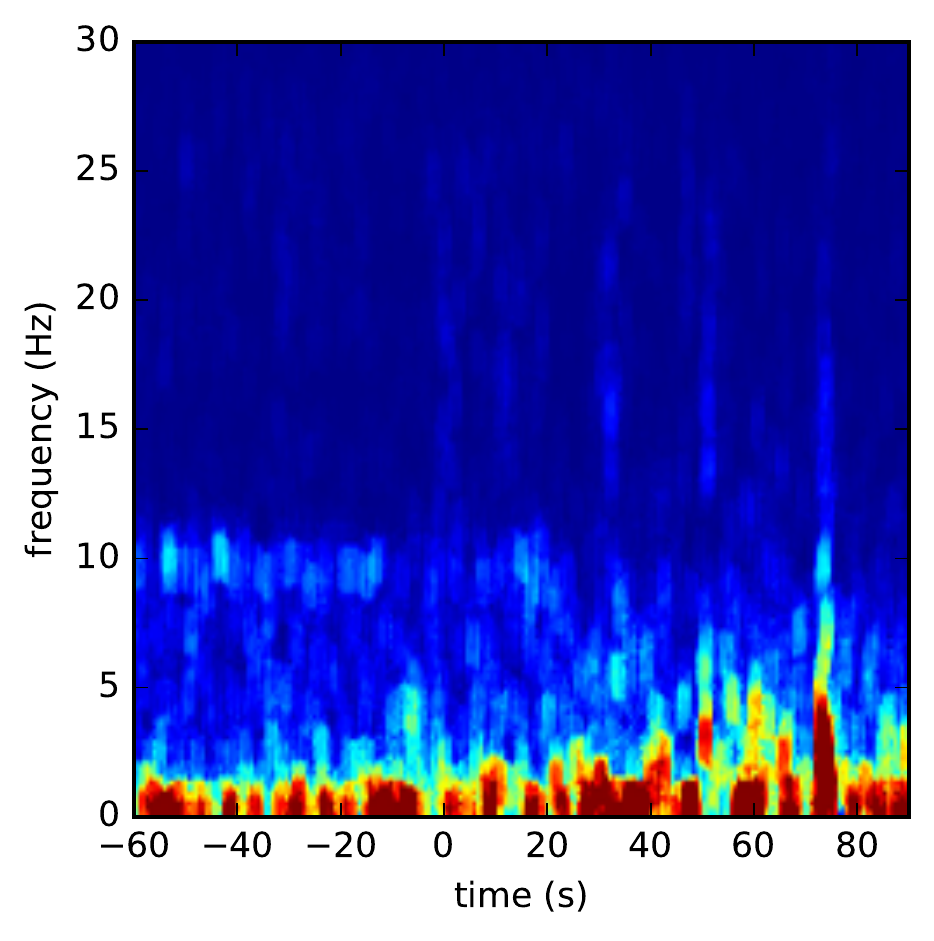}}
%  \vspace{1.5cm}
  \centerline{(b) Non-REM 1}\medskip
\end{minipage}
\hfill
\begin{minipage}[b]{0.19\linewidth}
  \centering
  \centerline{\includegraphics[width=\linewidth]{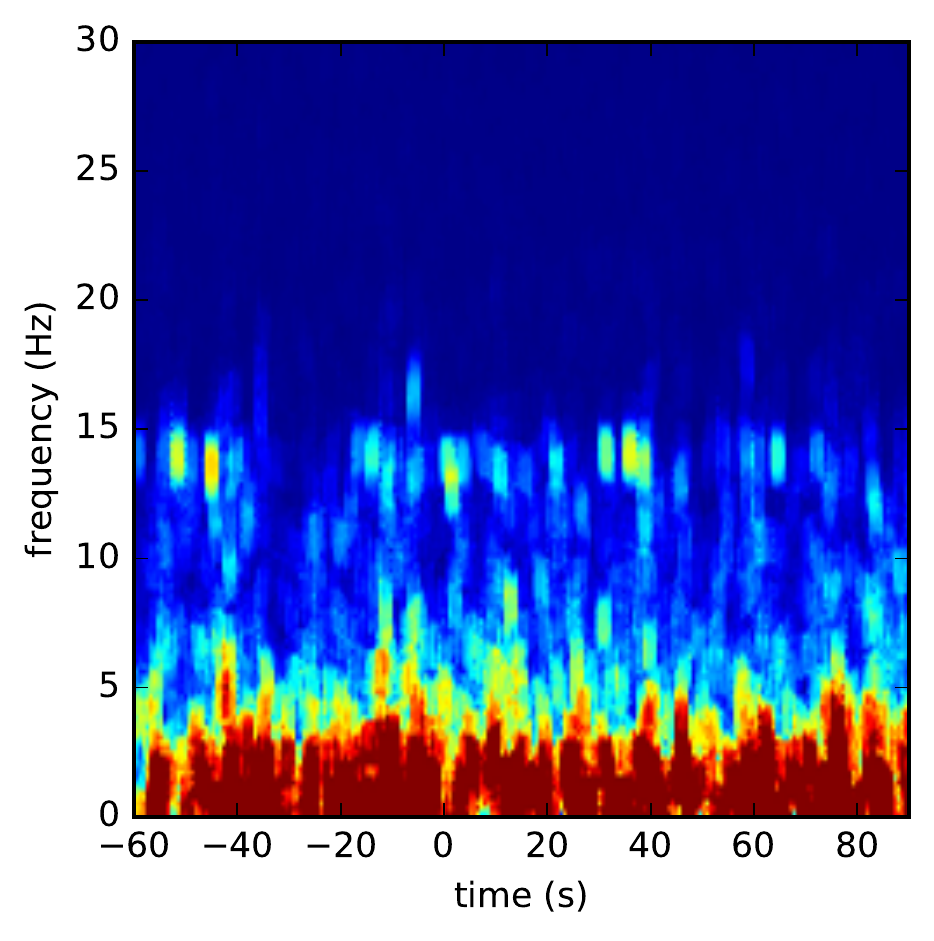}}
%  \vspace{1.5cm}
  \centerline{(c) Non-REM 2}\medskip
\end{minipage}
\hfill
\begin{minipage}[b]{0.19\linewidth}
  \centering
  \centerline{\includegraphics[width=\linewidth]{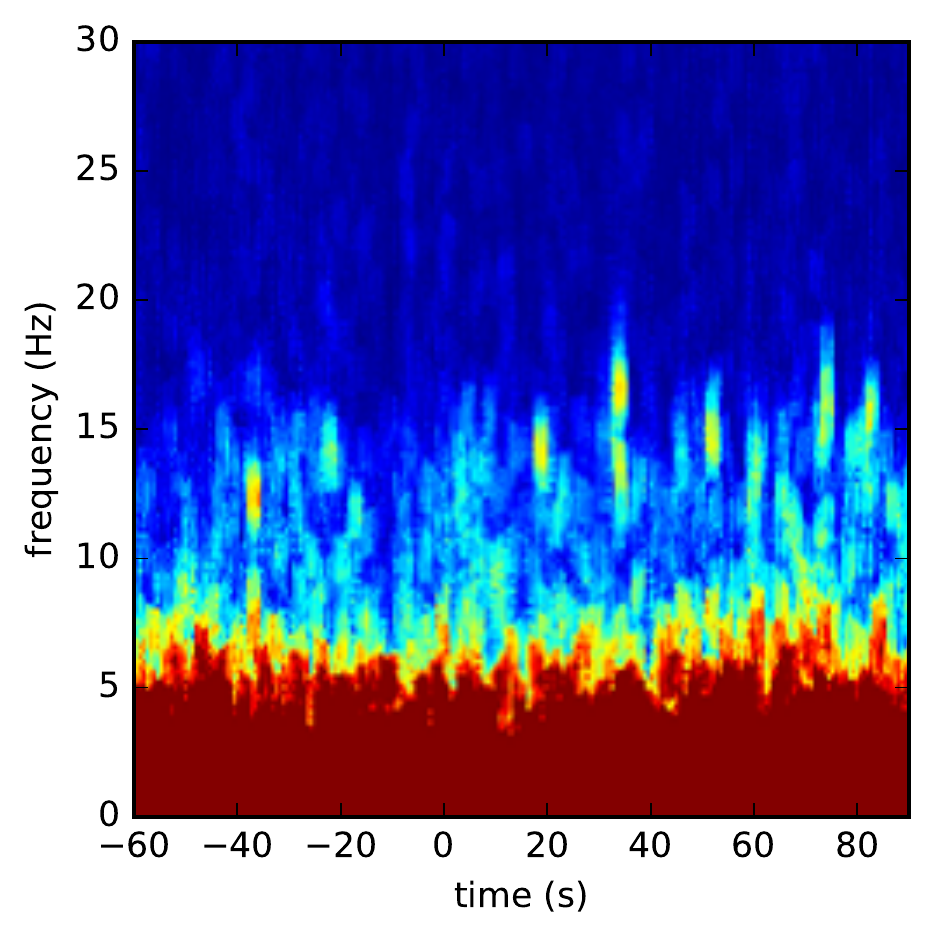}}
%  \vspace{1.5cm}
  \centerline{(d) Non-REM 3}\medskip
\end{minipage}
\hfill
\begin{minipage}[b]{0.19\linewidth}
  \centering
  \centerline{\includegraphics[width=\linewidth]{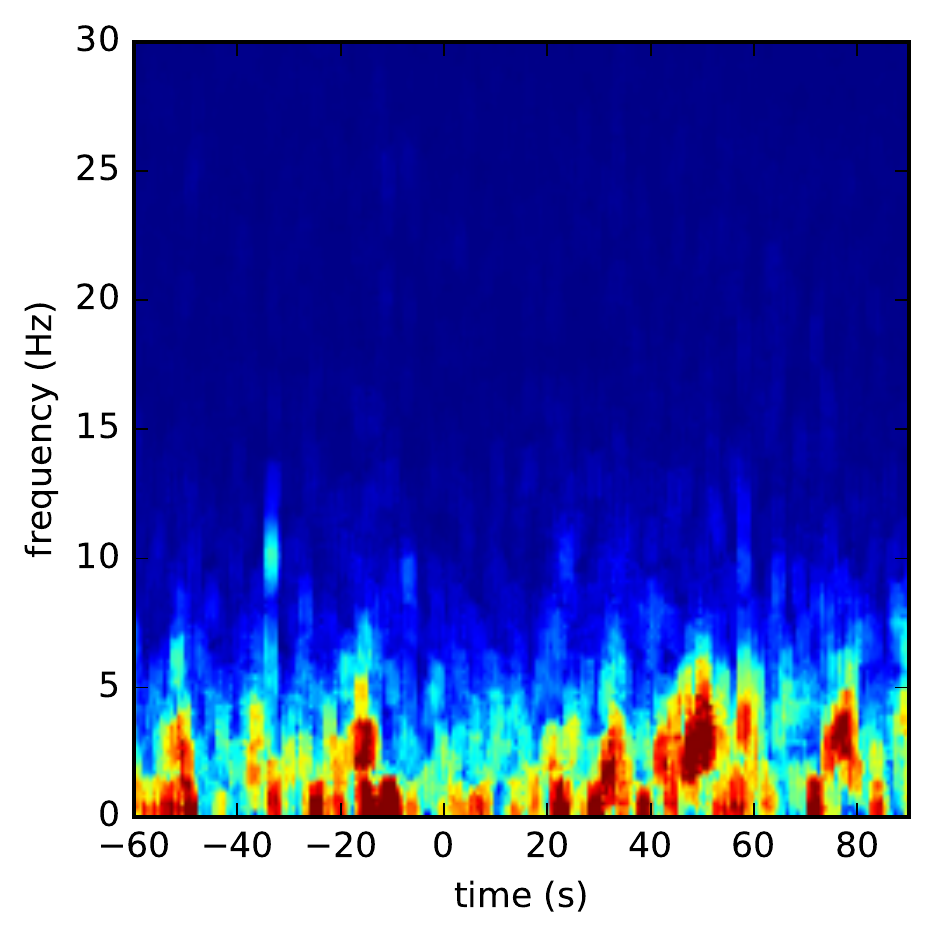}}
%  \vspace{1.5cm}
  \centerline{(e) REM}\medskip
\end{minipage}
\begin{minipage}[b]{0.19\linewidth}
  \centering
  \centerline{\includegraphics[width=\linewidth]{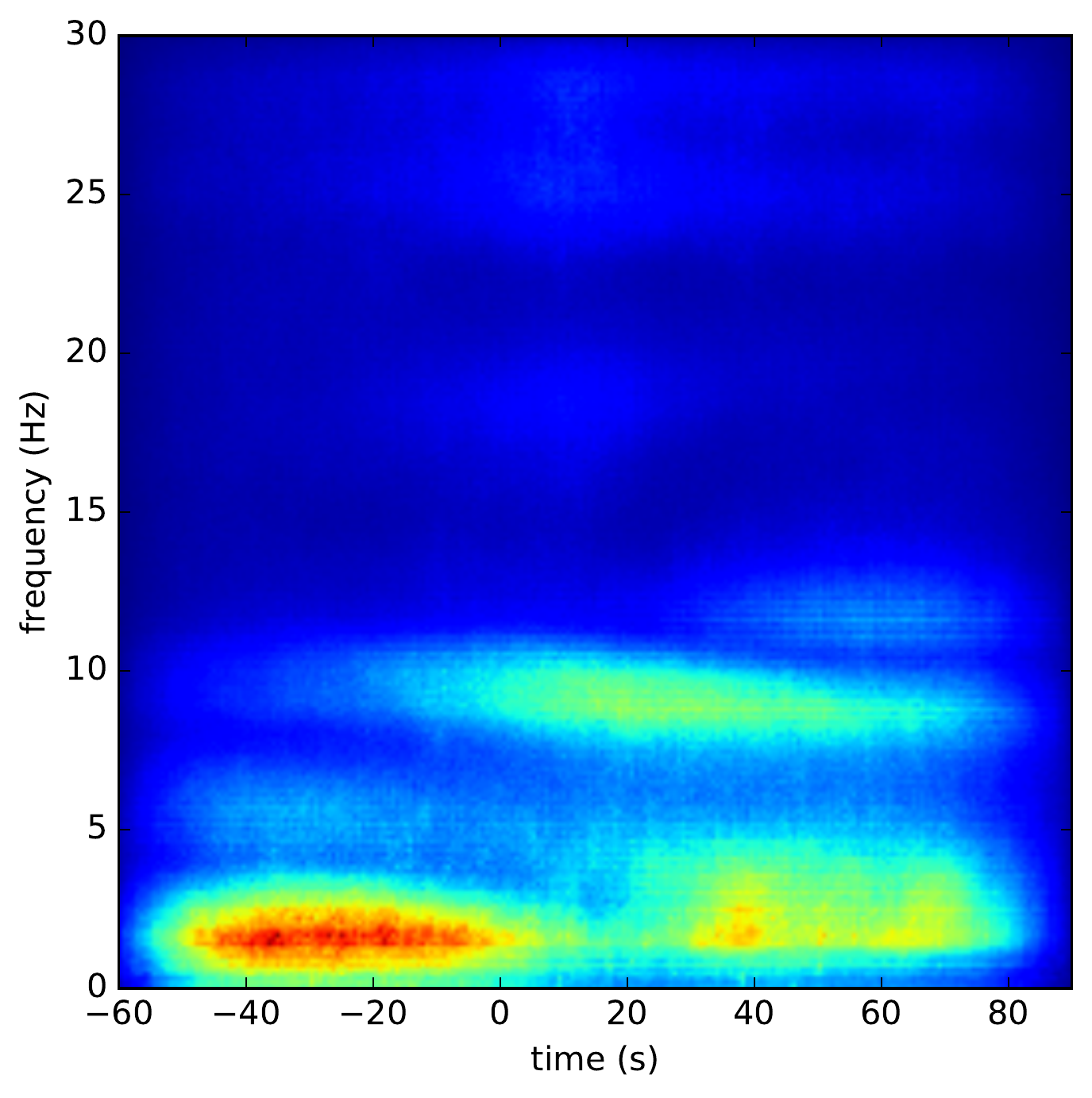}}
%  \vspace{2.0cm}
 % \centerline{(a) Wakefulness}\medskip
\end{minipage}
\hfill
\begin{minipage}[b]{0.19\linewidth}
  \centering
  \centerline{\includegraphics[width=\linewidth]{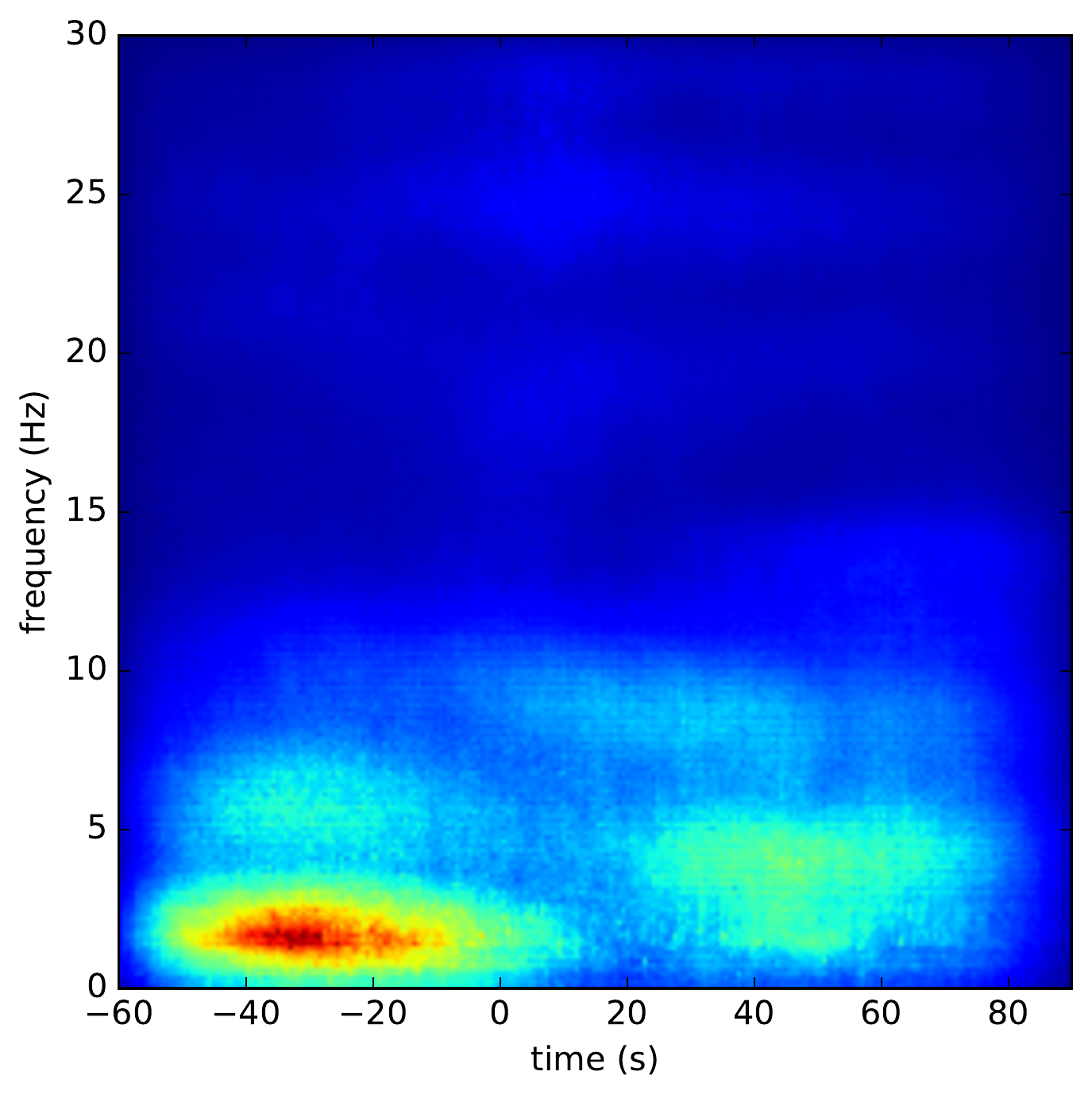}}
%  \vspace{1.5cm}
 % \centerline{(b) Non-REM 1}\medskip
\end{minipage}
\hfill
\begin{minipage}[b]{0.19\linewidth}
  \centering
  \centerline{\includegraphics[width=\linewidth]{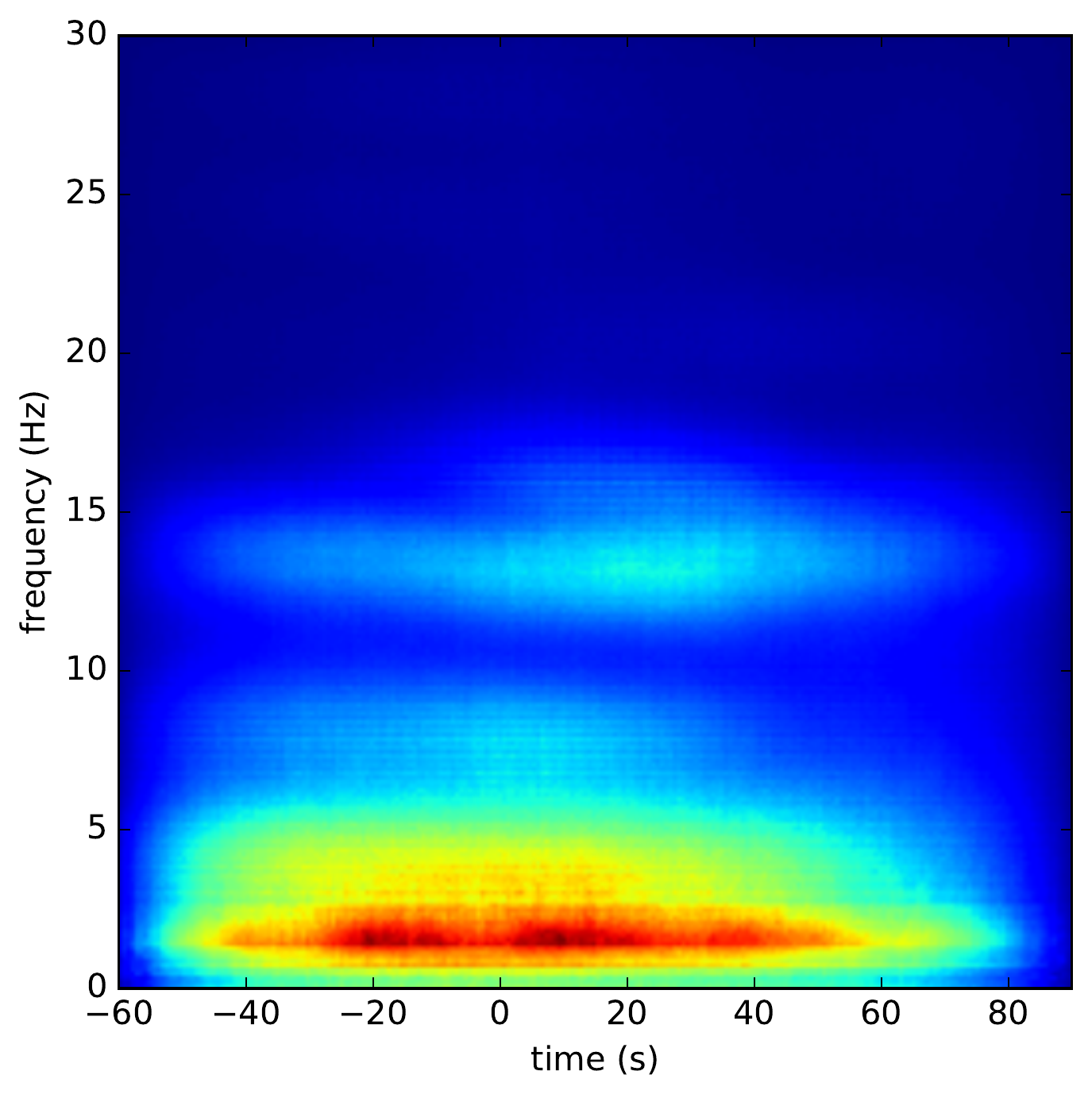}}
%  \vspace{1.5cm}
%  \centerline{(c) Non-REM 2}\medskip
\end{minipage}
\hfill
\begin{minipage}[b]{0.19\linewidth}
  \centering
  \centerline{\includegraphics[width=\linewidth]{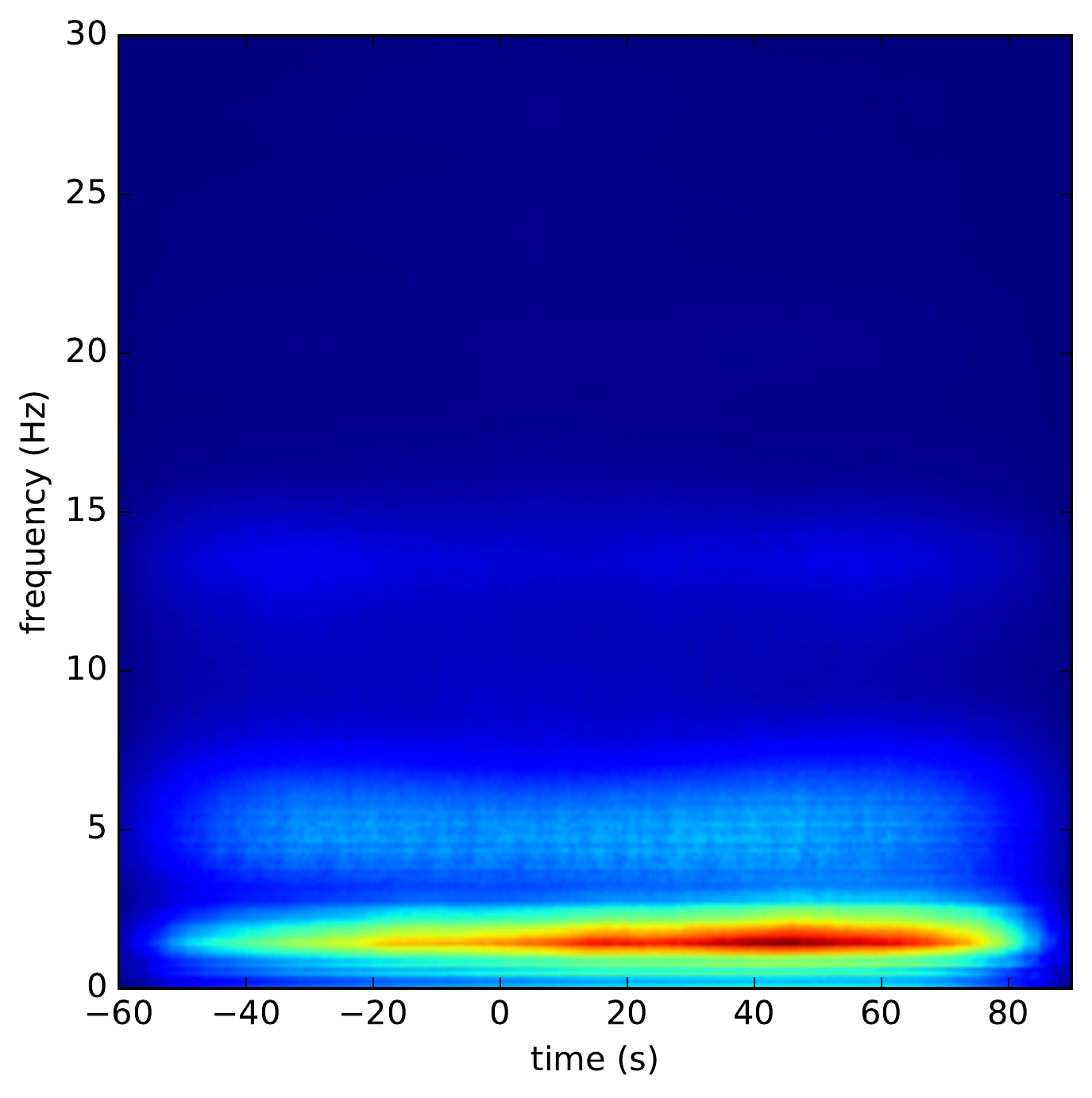}}
%  \vspace{1.5cm}
 % \centerline{(d) Non-REM 3}\medskip
\end{minipage}
\hfill
\begin{minipage}[b]{0.19\linewidth}
  \centering
  \centerline{\includegraphics[width=\linewidth]{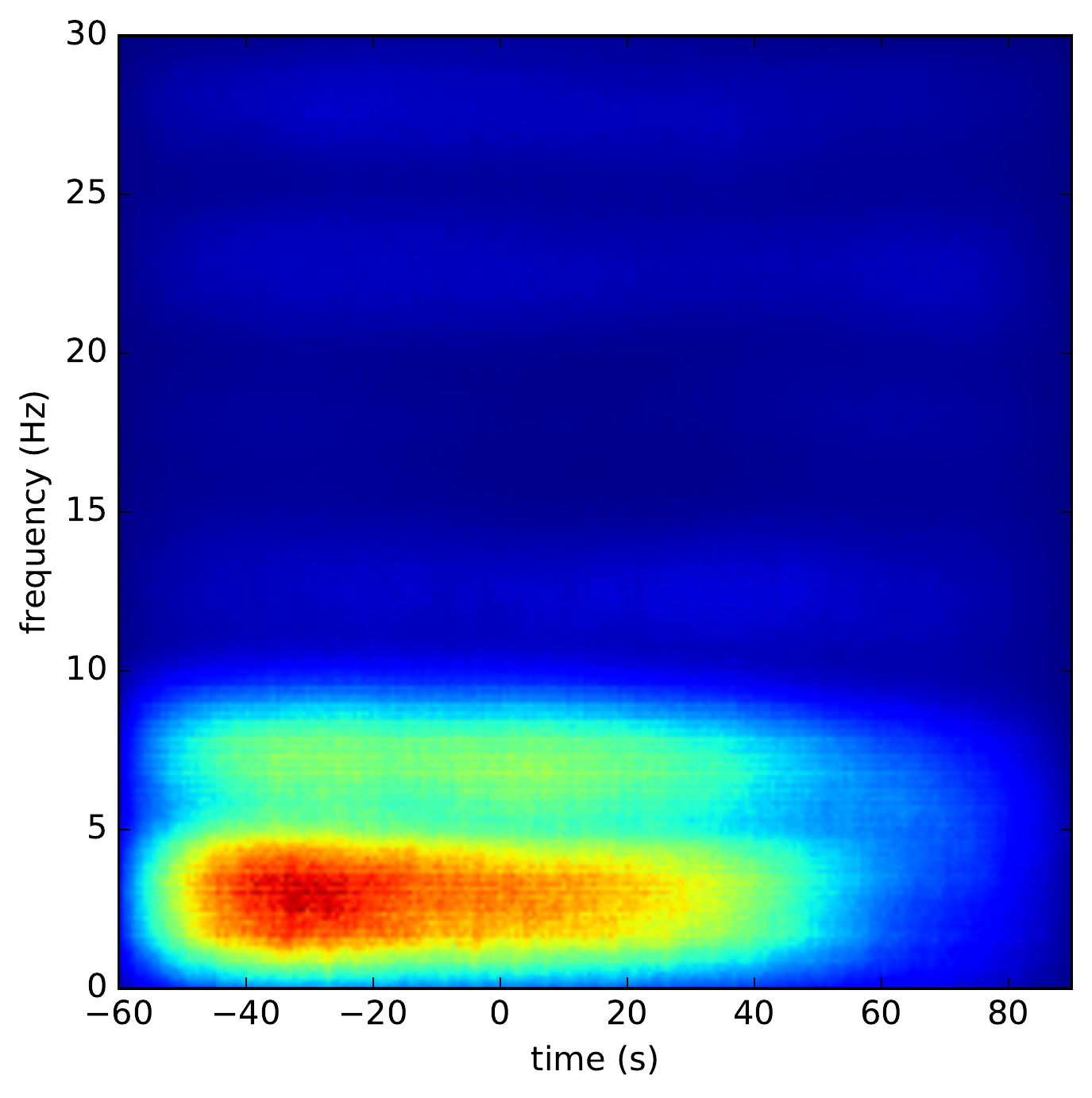}}
%  \vspace{1.5cm}
 % \centerline{(e) REM}\medskip
\end{minipage}
\vspace{-3mm}
\caption{Sensitivity analysis for subject 7. Top row shows characteristic sleep stage spectra; bottom row presents per-class sensitivity maps.}
\label{fig:class_sm}
\end{figure*}

Table~\ref{tab:confusion_matrix}, bottom block, shows the results obtained by VGG-FT. The improvement of this network with respect to VGG-FE is homogeneous across classes, the most prominent one being the increase of sensitivity for N1 class by $12\%$. This is achieved by reducing the misclassification between N1-W by $6\%$, N1-R by $5\%$ and N1-N2 by $1\%$. It is followed by an increase of sensitivity in R ($6\%$) and N2 ($4\%$), and precision in N1 ($4\%$), W ($3\%$) and R ($3\%$).

%Yet, this improvement comes mainly at the price of reducing W-class sensitivity, which wrongly places $20\%$ of the instances in the N1 class.

\begin{table}[htb]
\small
    \centering
  \begin{tabular}{c | c  c  c  c }
   % \toprule
    %\cmidrule{2-5}
    Study & Precision & Sensitivity & F$_1$-score & Accuracy\\
        %\midrule
    \hline
     \cite{Tsinalis2016a} & 92--\textbf{93}--94 & 75--\textbf{78}--80 & 82--\textbf{84}--86 & 84--\textbf{86}--88 \\
     \cite{Tsinalis2016b} & 90--\textbf{91}--92 & 71--\textbf{74}--76 & 79--\textbf{81}--83 & 80--\textbf{82}--84 \\
     %Rozhina?\\DeepSleep? & 0 & 0 & 0 & 0  \\
     VGG-FE & 90--\textbf{91}--93 & 70--\textbf{73}--77 & 78--\textbf{81}--83 & 81--\textbf{83}--85  \\     
     VGG-FT & 92--\textbf{93}--94 & 75--\textbf{78}--81 & 82--\textbf{84}--86 & 84--\textbf{86}--88  \\     
     \hline
    %\bottomrule
  \end{tabular}
  \caption{Comparison to existing literature. Mean values in bold within the corresponding 95\% confidence interval.}
  \label{tab:literature_comparison}
     \vspace{-2mm}
\end{table}

In Table~\ref{tab:literature_comparison}, we compare the performance of our method to the existing literature. As can be seen, both VGG-FE and VGG-FT exhibit state-of-the-art results, VGG-FT aligning with the best model to date \cite{Tsinalis2016a}.

\subsection{Visualisation}
\label{ssec:sleep_visualisation}
To better understand the internal representation of the network, we selected our best performing subject (i.e., subject 7) and calculated the sensitivity maps of each class independently. Fig.~\ref{fig:class_sm}, top row, depicts the most accurately classified spectrogram of each sleep stage, as characteristic examples of inputs to the classification system. Notice that the current epoch for classification is at the centre of the spectrogram (time between $0$ and $30$ s.), the two previous epochs span between $[-60,0]$ s. and the two posterior epochs are placed within the $[30,90]$ s. interval. Fig.~\ref{fig:class_sm}, bottom row, shows the per-class sensitivity map, each generated according to Eq.~\ref{eq:sensitivity} using only instances of subject 7 for this specific class, summing over the RGB colour channels, followed by 0-1 normalisation and converted back to RGB image using the `Jet' colour map. Now, we analyse them according to the sleep stage definitions in Section~\ref{sec:sleep}: Fig.~\ref{fig:class_sm}-a presents high sensitivity in the alpha band (8-12 Hz), characteristic of W stage, mainly centred at the current epoch and the immediately following one, maybe in order to be able to identify a transition stage. Sensitivity to low frequency activity corresponding to eye blinks is also evident. N1 stage (Fig.~\ref{fig:class_sm}-b) seems to show slightly decreased sensitivity in the alpha band as compared to W and rather higher sensitivity to theta (4-7 Hz) in the preceding epochs. Interestingly, Fig.~\ref{fig:class_sm}-c exhibits increased sensitivity around the upper sigma band (12-15 Hz) for the current epoch and low frequencies ($<$1.5 Hz) across the whole image, which might correspond to the network identifying spindles and K-complexes, respectively, which are characteristic features of the N2 stage. Slow wave activity (0.5-3 Hz) seems to be present in the N3 sensitivity map (Fig.~\ref{fig:class_sm}-d), with highest impact in the current and succeeding epochs. Finally, wide power band spanning from approximately 0.5 to 9 Hz is present in Fig.~\ref{fig:class_sm}-e, probably accounting for the mixed-frequency signal distinctive of R stage.

\section{Conclusions}
\label{sec:conclusions}
We have demonstrated that classification of sleep stages can be effectively framed as a visual task by first creating natural colour like images using multitaper spectral estimation and then applying recent achievements in the object recognition field to obtain state-of-the-art classification accuracy. Moreover, this approach greatly enhances the interaction with the domain expert by providing interpretable patterns to make sense of as well as a framework based on sensitivity analysis to easily inspect the network's reasoning. We think that the tools presented here can transcend EEG sleep scoring and be applied to other tasks within EEG analysis or, more generally, to other biological domains (e.g., EMG) where time-frequency signals are recorded. Further improvement of the method includes better hyperparameter optimisation when generating the spectral images. A thorough study of the obtained VGGNet layers might also be of interest to gain a deeper understanding of the internal structure of the network.

% Below is an example of how to insert images. Delete the ``\vspace'' line,
% uncomment the preceding line ``\centerline...'' and replace ``imageX.ps''
% with a suitable PostScript file name.
% -------------------------------------------------------------------------
%\begin{figure}[htb]
%
%\begin{minipage}[b]{1.0\linewidth}
%  \centering
%  \centerline{\includegraphics[width=8.5cm]{image1}}
%%  \vspace{2.0cm}
%  \centerline{(a) Result 1}\medskip
%\end{minipage}
%%
%\begin{minipage}[b]{.48\linewidth}
%  \centering
%  \centerline{\includegraphics[width=4.0cm]{image3}}
%%  \vspace{1.5cm}
%  \centerline{(b) Results 3}\medskip
%\end{minipage}
%\hfill
%\begin{minipage}[b]{0.48\linewidth}
%  \centering
%  \centerline{\includegraphics[width=4.0cm]{image4}}
%%  \vspace{1.5cm}
%  \centerline{(c) Result 4}\medskip
%\end{minipage}
%%
%\caption{Example of placing a figure with experimental results.}
%\label{fig:res}
%%
%\end{figure}

%\vfill
%\pagebreak

\begin{spacing}{0.965}
\bibliographystyle{IEEEbib}
\bibliography{references}

\begin{thebibliography}{10}

\bibitem{Mikkelsen2015}
K.B. Mikkelsen et~al.,
\newblock ``{EEG} recorded from the ear: Characterizing the ear-{EEG} method,''
\newblock {\em Frontiers in Neuroscience}, vol. 9, pp. 438, 2015.

\bibitem{Berry2017}
R.B. Berry et~al.,
\newblock {\em {The AASM Manual for the Scoring of Sleep and Associated Events:
  Rules, Terminology and Technical Specifications}},
\newblock American Academy of Sleep Medicine, 2017,
\newblock Version 2.4.

\bibitem{Aboalayon2016}
K.A.I. Aboalayon et~al.,
\newblock ``Sleep stage classification using {EEG} signal analysis: A
  comprehensive survey and new investigation,''
\newblock {\em Entropy}, vol. 18, no. 9, pp. 1--31, 2016.

\bibitem{Ronzhina2012}
M.~Ronzhina et~al.,
\newblock ``Sleep scoring using artificial neural networks,''
\newblock {\em Sleep Medicine Reviews}, vol. 16, no. 3, pp. 251 -- 263, 2012.

\bibitem{Ebrahimi2008}
F.~Ebrahimi et~al.,
\newblock ``Automatic sleep stage classification based on {EEG} signals by
  using neural networks and wavelet packet coefficients,''
\newblock in {\em 30th Annual International Conference of the IEEE Engineering
  in Medicine and Biology Society}, 2008, pp. 1151--1154.

\bibitem{Hsu2013}
Y-L. Hsu et~al.,
\newblock ``Automatic sleep stage recurrent neural classifier using energy
  features of {EEG} signals,''
\newblock {\em Neurocomputing}, vol. 104, pp. 105--114, 2013.

\bibitem{Supratak2017}
A.~Supratak et~al.,
\newblock ``{DeepSleepNet}: a model for automatic sleep stage scoring based on
  raw single-channel {EEG},''
\newblock {\em ArXiv e-prints}, vol. 1703.04046, 2017.

\bibitem{LeCun1989}
Y.~LeCun et~al.,
\newblock ``Backpropagation applied to handwritten zip code recognition,''
\newblock {\em Neural Computation}, vol. 1, no. 4, pp. 541--551, 1989.

\bibitem{Krizhevsky2012}
A.~Krizhevsky et~al.,
\newblock ``Imagenet classification with deep convolutional neural networks,''
\newblock in {\em Advances in Neural Information Processing Systems}, pp.
  1097--1105. 2012.

\bibitem{Simonyan2014}
K.~Simonyan and A.~Zisserman,
\newblock ``Very deep convolutional networks for large-scale image
  recognition,''
\newblock {\em CoRR}, vol. abs/1409.1556, 2014.

\bibitem{He2016}
K.~He et~al.,
\newblock ``Deep residual learning for image recognition,''
\newblock {\em CoRR}, vol. abs/1512.03385, 2015.

\bibitem{Tsinalis2016a}
O.~Tsinalis, P.M. Matthews, and Y.~Guo,
\newblock ``Automatic sleep stage scoring using time-frequency analysis and
  stacked sparse autoencoders,''
\newblock {\em Annals of Biomedical Engineering}, vol. 44, no. 5, pp.
  1587--1597, 2016.

\bibitem{Tsinalis2016b}
O.~Tsinalis et~al.,
\newblock ``Automatic sleep stage scoring with single-channel {EEG} using
  convolutional neural networks,''
\newblock {\em CoRR}, vol. abs/1610.01683, 2016.

\bibitem{Simonyan2013}
K.~Simonyan et~al.,
\newblock ``Deep inside convolutional networks: Visualising image
  classification models and saliency maps,''
\newblock {\em CoRR}, vol. abs/1312.6034, 2013.

\bibitem{Zeiler2014}
M.D. Zeiler and R.~Fergus,
\newblock ``Visualizing and understanding convolutional networks,''
\newblock {\em CoRR}, vol. abs/1311.2901, 2014.

\bibitem{Szegedy2015}
C.~Szegedy et~al.,
\newblock ``Going deeper with convolutions,''
\newblock {\em CoRR}, vol. abs/1409.4842, 2014.

\bibitem{Yosinski2014}
J.~Yosinski et~al.,
\newblock ``How transferable are features in deep neural networks?,''
\newblock {\em CoRR}, vol. abs/1411.1792, 2014.

\bibitem{Razavian2014}
A.S. Razavian et~al.,
\newblock ``{CNN} features off-the-shelf: an astounding baseline for
  recognition,''
\newblock {\em CoRR}, vol. abs/1403.6382, 2014.

\bibitem{Donahue2013}
J.~Donahue et~al.,
\newblock ``{DeCAF}: A deep convolutional activation feature for generic visual
  recognition,''
\newblock {\em CoRR}, vol. abs/1310.1531, 2013.

\bibitem{Thomson1982}
D.J. Thomson,
\newblock ``Spectrum estimation and harmonic analysis,''
\newblock {\em Proceedings of the IEEE}, vol. 70, no. 9, pp. 1055--1096, 1982.

\bibitem{Prerau2016}
M.J. Prerau et~al.,
\newblock ``Sleep neurophysiological dynamics through the lens of multitaper
  spectral analysis,''
\newblock {\em Physiology}, vol. 32, no. 1, pp. 60--92, 2016.

\bibitem{Glorot2010}
X.~Glorot and Y.~Bengio,
\newblock ``Understanding the difficulty of training deep feedforward neural
  networks,''
\newblock in {\em Proceedings of the Intl. Conf. on Artificial Intelligence and
  Statistics}, 2010.

\bibitem{Rasmussen2011}
P.~M. Rasmussen et~al.,
\newblock ``Visualization of nonlinear kernel models in neuroimaging by
  sensitivity maps,''
\newblock {\em NeuroImage}, vol. 55, no. 3, pp. 1120--1131, 2011.

\bibitem{Hashem1992}
S.~Hashem,
\newblock ``Sensitivity analysis for feedforward artificial neural networks
  with differentiable activation functions,''
\newblock in {\em Proceedings of the International Joint Conference on Neural
  Networks}, 1992, vol.~1, pp. 419--424.

\bibitem{Zurada1994}
J.M. Zurada et~al.,
\newblock ``Sensitivity analysis for minimization of input data dimension for
  feedforward neural network,''
\newblock in {\em Proceedings of the IEEE International Symposium on Circuits
  and Systems}, 1994, vol.~6, pp. 447--450.

\bibitem{Kemp2000}
B.~Kemp et~al.,
\newblock ``Analysis of a sleep-dependent neuronal feedback loop: the slow-wave
  microcontinuity of the {EEG},''
\newblock {\em IEEE Transactions on Biomedical Engineering}, vol. 47, no. 9,
  pp. 1185--1194, 2000.

\bibitem{Goldberger2000}
A.L. Goldberger et~al.,
\newblock ``Physiobank, physiotoolkit, and physionet,''
\newblock {\em Circulation}, vol. 101, no. 23, pp. e215--e220, 2000.

\bibitem{Srivastava2014}
N.~Srivastava et~al.,
\newblock ``Dropout: A simple way to prevent neural networks from
  overfitting,''
\newblock {\em Journal of Machine Learning Research}, vol. 15, pp. 1929--1958,
  2014.

\bibitem{Kingma2014}
D.P. Kingma and J.~Ba,
\newblock ``Adam: {A} method for stochastic optimization,''
\newblock {\em CoRR}, vol. abs/1412.6980, 2014.

\end{thebibliography}
\end{spacing}

\end{document}